\documentclass[10pt]{article}

\usepackage[letterpaper,margin=1in]{geometry}
\usepackage[T1]{fontenc}
\usepackage[utf8]{inputenc}
\usepackage{newtxtext,newtxmath}
\usepackage{microtype}
\usepackage{amsmath}
\usepackage{booktabs}
\usepackage[hidelinks]{hyperref}
\usepackage{float}
\usepackage{placeins}
\usepackage{needspace}
\usepackage[font=small,labelfont=bf,labelsep=period]{caption}
\usepackage[authoryear,round]{natbib}
\usepackage{flushend}

\newcommand{\repo}{\href{https://github.com/victorlavrenko/answer-engineering}{\nolinkurl{github.com/victorlavrenko/answer-engineering}}}
\newcommand{\release}{\href{https://github.com/victorlavrenko/answer-engineering/releases/tag/instruction-duplication-arxiv-v1}{\texttt{instruction-duplication-arxiv-v1}}}
\newcommand{\colab}{\href{https://colab.research.google.com/github/victorlavrenko/answer-engineering/blob/main/notebooks/instruction-placement-reproduction.ipynb}{\texttt{instruction-placement-reproduction.ipynb}}}

\hypersetup{
  pdftitle={Instruction Duplication as an Inference-Time Control Primitive},
  pdfauthor={Victor Lavrenko},
  pdfsubject={Inference-time control, instruction following, trajectory editing, Answer Engineering}
}

\title{\vspace{-1.2em}\textbf{Instruction Duplication as an Inference-Time Control Primitive}}
\author{Victor Lavrenko\\
\small PeaceTech VC, Israel\\
\small \href{mailto:victor@peacetech.vc}{victor@peacetech.vc}}
\date{3 September 2026}

\begin{document}

\twocolumn[
\begin{@twocolumnfalse}
\maketitle
\vspace{-1.1em}
\begin{abstract}
Procedural instruction following is a basic requirement for controllable language-model systems, especially when generated trajectories are inspected or repaired downstream. We introduce \emph{instruction duplication}, a minimal black-box inference-time control that repeats only the procedural instruction, without retraining or decoding changes. Across seven instruction-tuned models, 300 medical multiple-choice questions, eight placement conditions, and 16,800 scheduled generations, moving from one to two copies raises the deterministic All-8 diagnostic---responses passing all eight observable tests---from 90.22\% to 93.17\% (+2.95 percentage points), eliminating 30.2\% of the failures remaining after one copy. Pre-provisional TF--IDF recall rises from 73.44\% to 74.81\% (+1.38 points; Holm-adjusted $p<.001$), while final-answer accuracy remains exactly 60.21\%. Premature commitment increases from 1.52\% to 2.30\% ($p_{\rm Holm}=.00536$). A blinded challenge audit yields 10/30 directional confirmations, 20/30 perceptual ties, and no reversals; its prespecified 28/30 confirmation criterion is not met. Yet this distinction can matter operationally when a downstream system acts on the generated trajectory. In Answer Engineering (AE), where explicit trajectory state determines local repair, the published reason-first no-editing SSNHL endpoint was 25.1\%; system-only AE was later reproduced at 84.2\%, and the same trailing duplicate raised it to 97.1\%. For conductive diagnostic branch preservation, the corresponding values are 58.9\% published without editing, 78.6\% with reproduced AE, and 73.8\% with AE plus duplication---a within-AE decrease, but still 14.9 points above the no-editing baseline. Instruction duplication is therefore a low-complexity, placement-sensitive control whose practical value can emerge through the downstream system that consumes the exposed trajectory.
\end{abstract}
\vspace{0.8em}
\end{@twocolumnfalse}
]

\section{Introduction}
Language-model systems often need more than a final answer. A verifier, auditor, or runtime controller may require the model to inventory evidence, compare alternatives, delay commitment, expose a decisive distinction, or reconsider an intermediate decision. Making these operations explicit creates machine-addressable states at which downstream mechanisms can detect or repair failures. Procedural state and task correctness are distinct: a response can be correct while omitting a state that a controller needs, or can execute a protocol while reaching the wrong answer.

\citet{leviathan2025prompt} show that apparently redundant repetition can be a real inference-time intervention: whole-prompt repetition transforms $\langle Q\rangle$ into $\langle Q\rangle\langle Q\rangle$ and improves non-reasoning accuracy across model families. \citet{xu2023rereading} likewise show that re-reading the question can improve reasoning. Instruction position also affects sequence generation, motivating after-input placement \citep{liu2024position}. We ask a narrower control question: if the substantive query is unchanged, can repeating only the \emph{procedural instruction} change the explicit state available to a downstream system?

\Needspace{4\baselineskip}
Let $I$ denote a fixed instruction and $Q$ the substantive query. A conventional prompt may contain one copy, $\langle I\rangle\langle Q\rangle$; an instruction-duplicated prompt may contain two, for example $\langle I\rangle\langle Q\rangle\langle I\rangle$. The extra copy adds no new rule, example, reward, verifier, or model access. It changes only how often and where the same procedural instruction is exposed before generation. Unlike whole-prompt repetition, the query itself is not duplicated; the intervention targets execution of the requested procedure rather than re-exposure to task content.

We make four contributions. First, we define instruction duplication as a minimal black-box control mechanism. Second, a complete $2\times2\times2$ placement factorial separates copy count from location. Third, the experiment shows a control-relevant dissociation: explicit protocol state changes while aggregate answer accuracy does not. Fourth, a blinded challenge audit and an AE downstream experiment show why that dissociation can matter: a change that is small or even perceptually negligible to a reader can alter the explicit state on which a deterministic trajectory editor operates. The present design illustrates this systems interaction without claiming that any single primary metric mediates the AE effect.

\Needspace{7\baselineskip}
\section{Related Work}
Instruction tuning and preference optimization improve broad responsiveness by changing model parameters \citep{ouyang2022training}. \citet{lightman2024verify} instead provide learning signals over intermediate steps through process supervision. At inference time, \citet{wei2022cot} show that chain-of-thought prompting and decomposition can change the visible trajectory without retraining, although visible rationales need not faithfully reveal the internal causes of an answer \citep{turpin2023unfaithful}. More direct control mechanisms include grammar-constrained decoding \citep{geng2023grammar} and instruction-specific activation steering, which can improve constraint following \citep{stolfo2025steering}. Instruction duplication lies at the black-box end of this spectrum: it adds only prompt tokens and requires no model internals.

Clinical diagnostic reasoning provides a domain-grounded rationale for the contrastive protocol. \citet{kassirer1983hypothesis} describes diagnosis as iterative hypothesis testing: clinicians generate candidate diagnoses, compare them with findings, rule alternatives in or out, and revise the hypotheses that survive. Illness-script accounts from \citet{charlin2000scripts} likewise emphasize plausible hypotheses followed by comparison against discriminating signs and symptoms. More recently, \citet{mcduff2025ddx} treat differential diagnosis with LLMs as construction and refinement of a ranked differential rather than only a one-shot label. Our protocol expands this normally compressed process so that requested operations can be inspected mechanically.

Our evaluation also relates to work separating verifiable instruction compliance from broader semantic quality \citep{zhou2023ifeval,lou2023survey}. Unlike whole-prompt or question repetition, we repeat only the procedural instruction and measure pre-commitment state. Unlike after-input training arrangements, the full placement factorial is an inference-time intervention. \citet{lavrenko2026ae} introduced AE, a trajectory editor that can act only on information and intermediate decisions that become observable; our endpoint is that observable substrate.

\section{Experiment}
\subsection{Models, Questions, and Factorial Conditions}
We evaluate seven visible-output instruction-tuned models: Gemma 3 12B, Llama 3.3 70B Instruct, Llama 4 Scout, Ministral 3 14B Instruct, Mistral Large 3, Qwen3 30B-A3B Instruct, and Qwen3 235B-A22B Instruct. The benchmark contains 300 medical multiple-choice questions, 100 each from MedQA, MedXpertQA, and AfriMed-QA \citep{jin2021medqa,zuo2025medxpertqa,nimo2025afrimed}. Medical stems frequently contain measurements, negations, timing qualifiers, and plausible alternatives.

The same instruction may appear in the system message ($S$), before the question ($B$), and/or after the question ($A$). The $2\times2\times2$ design therefore contains zero copies, three one-copy conditions, three two-copy conditions, and one three-copy condition: $7\times300\times8=16{,}800$ scheduled cells. The frozen run contains 16,646 completed generations, 152 truncations, and 2 hard failures. Generation used temperature 0, cell-specific seeds, pinned model/provider routes, and model-specific output ceilings.

\subsection{Procedural Instruction}
The instruction asks the model to use eight headings in order and avoid selecting an answer before the provisional-answer section. It deliberately operationalizes a hypothetico-deductive, contrastive form of diagnostic reasoning: the model should treat the leading diagnosis as a best-supported working hypothesis, compare it with a strong competitor, identify discriminating evidence, ask what would change the ranking, and reconsider before committing \citep{kassirer1983hypothesis,charlin2000scripts,bowen2006reasoning}. The eight requested operations are: (1) Facts; (2) Implications; (3) Provisional answer; (4) Best alternative; (5) Decisive distinction; (6) What would change the answer; (7) Reconsideration; and (8) Final answer.

\subsection{Measurements and Eligibility}
We keep six exploratory outcomes conceptually separate and Holm-correct their pooled one-copy versus two-copy tests as one family. \textbf{Section completion} counts identifiable requested sections with non-trivial generated content (0--8). \textbf{Deterministic role completion} applies deterministic content tests to seven roles; the sixth ``what would change'' role uses the observable criterion that its section is identifiable and non-trivial. \textbf{Contrastive discussion} summarizes completion of provisional answer, distinct alternative, decisive distinction, what-would-change, and reconsideration. \textbf{Premature commitment} marks firm answer selection before the provisional role and is evaluated among completed generations. \textbf{Accuracy} is the final multiple-choice answer under intention-to-treat (ITT). Generation truncation and hard failure are reported separately rather than folded into premature commitment.

\textbf{Pre-provisional TF--IDF recall} is the fraction of the question stem's TF--IDF-weighted lexical content recovered in the generated Facts and Implications sections before provisional-answer selection. The denominator is the total sublinear TF--IDF weight of non-stopword stem terms; the numerator credits matching terms exposed in those sections, capped at their stem frequency. It measures lexical exposure rather than factual completeness or semantic understanding. IDF weights use a frozen global document-frequency table over 15,103,887 PubMed abstracts rather than frequencies estimated from the 300-question sample \citep{salton1988term}.

TF--IDF is eligible for 257/300 questions. Eligibility is fixed once at the question level from the frozen source inventory, independently of response or condition. A stem must contain at least one standalone factual statement outside its interrogative tail. The 43 exclusions are topic-only or choice-dependent stems without such a fact inventory (32), plus 11 that also trigger a very-short-stem flag. The same decision is reused for every model and all eight conditions.

\Needspace{7\baselineskip}
\subsection{Contrasts, Robustness Analyses, and\texorpdfstring{\\}{ }Multiplicity}
The pooled copy-count estimand averages the three two-copy conditions ($SB,SA,BA$) minus the three one-copy conditions ($S,B,A$) within each model-question block. We additionally report all eight condition means, the full balanced factorial main effects and interactions, a one-copy after-query contrast ($A$ versus the mean of $S,B$), and a trailing-copy contrast that adds an after-query copy to a conventional instruction ($S\rightarrow SA$ and $B\rightarrow BA$).

To test whether higher-copy cells are more than additive combinations of single locations, we estimate an additive model from the zero- and one-copy cells and test observed-minus-predicted residuals for each higher-copy cell by question-cluster sign flips. For protocol-dependent outcomes the zero-copy cell is coded 0 because the requested protocol is absent. Because these outcomes are bounded and near ceiling after one instruction, the residual test is a diagnostic for super-additivity rather than a linear mechanistic model.

Because duplication changes output length, we perform a post-hoc length robustness analysis on eligible one- and two-copy cells: within question-model fixed effects, TF--IDF recall is regressed on a two-copy indicator and pre-answer content-token count, with question-cluster robust standard errors. The same model is applied to the targeted trailing-copy pairs with a fixed effect for each question-model-pair. These regressions ask whether a condition difference remains among traces of comparable observed length.

Confidence intervals for paired contrasts use 10,000 question-cluster bootstrap resamples and two-sided $p$-values use 50,000 paired sign-flip draws. The six pooled outcomes above are Holm-corrected together \citep{holm1979}. All-8 completion is a separately reported derived conjunction with a cluster-bootstrap interval and nominal sign-flip test; it is not added post hoc to that Holm family. Model-specific diagnostics are Holm-corrected across seven models within endpoint.

\subsection{Blinded Human Challenge Audit}
The formal All-8 diagnostic is a conjunction of thresholded textual decisions, so we challenged the positive transitions that create its copy-count effect. The matching is exact: within the same question and model, each one-copy condition is paired only with a two-copy condition obtained by adding one location ($S\rightarrow SB/SA$, $B\rightarrow SB/BA$, $A\rightarrow SA/BA$). Questions used in prior judge/audit development were excluded. From the remaining machine $0\rightarrow1$ All-8 transitions, the tool selected 30 cases by stable-hash sampling proportional to the atomic changed-role patterns. Treatment identity, model, placement, automatic preference, stratum, and gold answer were hidden; A/B orientation was randomized. One non-clinician author judged only whether the displayed procedural criterion was visibly satisfied, not whether the medical answer was correct. Three degradation and three tie sentinels were added descriptively.

\section{Results}
\subsection{All Eight Conditions and\texorpdfstring{\\}{ }Factorial Decomposition}
Table~\ref{tab:conditions} reports every placement condition. One instruction already moves the bounded protocol measures close to their ceilings. Duplication then yields a smaller positive gain: mean TF--IDF rises from 73.44\% with one copy to 74.81\% with two (+1.38 points). Accordingly, the factorial decomposition shows strong saturation rather than additive gains: the first placement captures most of the available improvement and additional locations have diminishing returns (all TF--IDF location and interaction terms, sign-flip $p<.001$). The same pattern occurs for section, role, contrastive, and All-8 measures. Accuracy differs: none of its seven factorial terms is significant (all $p\ge .064$).

The explicit additivity test finds no positive super-additivity. Residuals are strongly negative for TF--IDF and the other bounded protocol measures ($p<.001$), consistent with saturation after the first instruction; accuracy residuals are null ($p=.30$--$.77$). Additional-copy effects are therefore incremental and placement-specific rather than synergistic on the raw scales.

At one copy, after-query placement alone is not a strong lexical lever: relative to the mean of system-only and before-only, $A$ changes TF--IDF by +0.31 points (95\% CI $[-0.19,0.83]$, $p=.239$) and accuracy by +0.62 points ($p=.392$). Section count rises by 0.040 ($p=.008$), while the other protocol outcomes are small or null. This rules out interpreting the present results as a general ``put the instruction last'' effect.

\begin{table*}[t]
\centering
\small
\setlength{\tabcolsep}{5.4pt}
\begin{tabular}{@{}lrrrrrrrr@{}}
\toprule
Condition & Copies & Sections & TF--IDF & Contrast & Roles & All 8 & Accuracy & Trunc.\\
\midrule
None & 0 & -- & -- & -- & -- & -- & 60.90 & 28/2100\\
System ($S$) & 1 & 7.939 & 72.35 & 96.67 & 7.796 & 88.81 & 60.38 & 16/2100\\
Before ($B$) & 1 & 7.909 & 74.32 & 97.57 & 7.831 & 92.57 & 59.62 & 24/2100\\
After ($A$) & 1 & 7.963 & 73.65 & 96.98 & 7.827 & 89.29 & 60.62 & 8/2100\\
$S+B$ & 2 & 7.931 & 73.39 & 97.45 & 7.828 & 91.52 & 60.33 & 18/2100\\
$S+A$ & 2 & 7.924 & 74.83 & 98.15 & 7.866 & 94.52 & 60.62 & 20/2100\\
$B+A$ & 2 & 7.911 & 76.22 & 97.68 & 7.840 & 93.48 & 59.67 & 21/2100\\
$S+B+A$ & 3 & 7.930 & 75.09 & 97.74 & 7.848 & 93.24 & 60.14 & 17/2100\\
\bottomrule
\end{tabular}
\caption{All eight conditions. TF--IDF, Contrast, All 8, and Accuracy are percentages; Sections and Roles are counts out of 8. Protocol-dependent metrics are not reported for the no-instruction cell; only the factorial/additivity diagnostic codes their absence as zero.}
\label{tab:conditions}
\end{table*}
\FloatBarrier

\subsection{Pooled One-to-Two Copy Effects and Length}
The pooled contrast is shown in Table~\ref{tab:main}. The deterministic All-8 diagnostic rises from 90.22\% to 93.17\% (+2.95 points; 95\% CI $[2.14,3.79]$), reducing machine-detected failures from 9.78\% to 6.83\%---a 30.2\% relative reduction. TF--IDF recall rises by +1.38 points and survives Holm correction, as do deterministically scored contrastive discussion and role count. Premature commitment increases by +0.77 points, a real adverse effect.

Accuracy is 3793/6300 (60.21\%) in both arms. This equality is a dissociation, not identical output: 341/2100 model--question blocks change their number of correct placements (170 favor two copies, 171 favor one). Separately, 480 blocks change their number of All-8-complete placements; 379/480 (79.0\%) do so with no accuracy-count change. The intervention changes the visible control surface without improving the pooled task endpoint.

\begin{table}[H]
\centering
\small
\setlength{\tabcolsep}{3.0pt}
\begin{tabular}{@{}lrrrr@{}}
\toprule
Outcome & 1 copy & 2 copies & $\Delta$ & $p_{\rm Holm}$\\
\midrule
\textbf{All 8 diagnostic (\%)}$^{b}$ & \textbf{90.22} & \textbf{93.17} & \textbf{+2.95} & --\\
Sections (0--8) & 7.937 & 7.922 & -0.015 & .244\\
TF--IDF recall (\%) & 73.44 & 74.81 & +1.38 & .00012\\
Contrastive (\%) & 97.07 & 97.76 & +0.69 & .00020\\
Roles (0--8) & 7.818 & 7.845 & +0.027 & .0356\\
Premature commit. (\%)$^{a}$ & 1.52 & 2.30 & +0.77 & .00536\\
Accuracy, ITT (\%) & 60.21 & 60.21 & +0.00 & 1.000\\
\bottomrule
\end{tabular}
\caption{Pooled two-copy minus one-copy effects. The six non-bold outcomes form one exploratory Holm family. $^{a}$Among completed generations. $^{b}$Derived conjunction reported separately; 95\% CI for $\Delta$ $[2.14,3.79]$, nominal sign-flip $p<.0001$.}
\label{tab:main}
\end{table}

Pre-answer content length increases from 201.0 to 216.0 tokens (+7.5\%). In the observed-length robustness regression, the pooled two-copy coefficient is +0.36 points (95\% CI $[-0.01,0.72]$, $p=.054$).

A trailing duplicate is the strongest placement-specific intervention. Comparing $S\rightarrow SA$ and $B\rightarrow BA$, TF--IDF recall rises from 73.33\% to 75.53\% (+2.19 points, 95\% CI $[1.76,2.64]$, $p<.0001$). The observed-length coefficient remains +1.24 points (95\% CI $[0.76,1.72]$, $p=4.5\times10^{-7}$). This contrast also raises All-8 by +3.31 points, contrastive discussion by +0.80 points, and roles by +0.040; section count and accuracy are unchanged. TF--IDF and All-8 point estimates are positive in all seven models; five and three models, respectively, survive within-endpoint Holm correction, while no model has a corrected accuracy effect. The pooled effect is therefore distributed rather than driven by one outlier.

Truncations by copy count 0/1/2/3 are 28/48/59/17; the pooled two-minus-one difference is +0.175 points (95\% CI $[-0.048,0.397]$, $p=.166$). Hard failures are 0/6300 versus 2/6300 ($p=.502$). Premature commitment, however, is higher in each two-copy placement than in the one-copy range and reaches 3.12\% with three copies.

\subsection{Deterministic All-8 Completion and Challenge Audit}
Of 30 machine-positive All-8 transitions in the blinded challenge audit, 20 are perceptual ties. All 10 non-tied pairs favor the duplicated/machine-preferred response (one-sided exact sign $p=.00098$), so the prespecified 28/30 confirmation criterion is not met. The audit therefore does not validate every automatic transition. Instead, it illustrates a narrower point: human-semantic and controller-operational equivalence can differ because a deterministic system may require an explicit alternative, qualifier, or retain/revise marker that a reader can infer from context.

Positive pooled role changes concentrate in provisional-answer completion, best-alternative handling, and reconsideration; the remaining roles are near ceiling.

\subsection{Operational Reading of the Effect}
The effect is best read as a change in \emph{available control state}, not as a uniform gain in answer quality. Accuracy is unchanged, while 379 of the 480 model--question blocks whose All-8 completion count changes do so without any accuracy-count change. The intervention can therefore reorganize what is exposed before commitment while leaving the pooled task endpoint fixed.

The placement pattern also rejects a simple ``more instruction is better'' rule. Most bounded protocol measures are already near ceiling after one copy, the additivity analysis is consistent with saturation, and the trailing-copy contrast is stronger than after-query placement alone. Meanwhile premature commitment rises and reaches 3.12\% with three copies. Copy count and location are therefore a small control surface with benefits and costs, not a monotone verbosity knob.

The audit marks the human/controller boundary: 20/30 machine-positive transitions are perceptual ties, yet all ten non-tied pairs favor the machine-preferred response. A human-facing system should not treat these threshold crossings as demonstrated quality gains. A parser, monitor, or editor may nevertheless care because a small explicit marker can determine whether a downstream action is available at all.

\Needspace{11\baselineskip}
\section{Discussion}
\subsection{What the Intervention Supports}
Instruction duplication changes explicit pre-commitment state. It removes 30.2\% of the machine-detected All-8 failures remaining after one instruction, increases question-specific lexical exposure, and does so without moving aggregate accuracy. The block-level dissociation shows that these are not two labels for the same behavioral change. Copy count and placement are independent controls over the exposed trajectory.

The factorial analysis is consistent with saturation rather than positive super-additivity: the first instruction supplies most of the bounded protocol effect, while subsequent copies have smaller placement-dependent consequences. The relevant control parameter is therefore copy count together with location.

\subsection{Why Trajectory Compliance Matters to Deterministic Controllers}
Formal compliance and human-perceived compliance are different endpoints. A human reader can often infer an omitted state from context; a deterministic controller generally cannot. A monitor cannot flag a qualifier that never appears in the visible state, and a local controller has no direct edit target if the model never names an alternative or never marks retain/revise status. We therefore interpret the procedural metrics as measures of explicit, machine-addressable trajectory state rather than as calibrated measures of human-perceived reasoning quality.

This is precisely where a modest improvement in instruction-following can become operationally important. \citet{lavrenko2026ae} describe Answer Engineering (AE), which mechanically detects explicit protocol state and locally edits the visible trajectory. If an instruction-placement change makes the generated trajectory more closely instantiate the requested procedure, the controller may encounter a different set of recognizable states and repair opportunities even when aggregate final-answer accuracy of ordinary generation does not move.

The original AE benchmark makes the progression explicit. With a reason-first prompt but \emph{without} trajectory editing, the published endpoint was 25.1\% on 1,000 SSNHL target cases; adding trajectory editing raised it to 83.5\%. In the later instruction-placement reproduction, system-only AE closely reproduced that edited result at 842/1000 (84.2\%), and adding the same trailing duplicate raised it further to 971/1000 (97.1\%, +12.9 points; approximate 95\% CI $[10.4,15.4]$; exact McNemar $p\approx7.2\times10^{-24}$). The clinical target is appropriate steroid treatment, but the frozen operational endpoint is deliberately narrower: the response must contain the substring ``steroid.'' Dose, route, urgency, and appropriateness are not separately scored.

The conductive scope clarifies the contrast rather than undermining it. In the published reason-first condition without trajectory editing, diagnostic branch preservation was 58.9\%; published AE raised it to 77.9\%, and the later system-only reproduction was 786/1000 (78.6\%). With duplication it fell to 738/1000 (73.8\%). Thus duplication reduced the marginal conductive benefit of AE, but the duplicated controller-equipped system still remained 14.9 points above the published no-editing baseline. Because the AE rules were designed around SSNHL failures and no conductive-specific treatment rules were added, a plausible interpretation is that duplication changed the initial trajectory distribution and left fewer locally recognizable conductive trajectories for the diagnosis-symmetric rules to improve.

\Needspace{6\baselineskip}
For this conductive contrast, success requires ``conductive'' and neither ``senso'' nor ``ssnhl.'' It does \emph{not} require steroid absence or score conductive-treatment correctness; it only tests whether the response remains on the conductive rather than SSNHL diagnostic branch. The AE evidence therefore illustrates the systems claim: duplication can matter because it changes the trajectory on which orchestration operates. The present experiments do not establish that TF--IDF, All-8, or any other single measured feature mediates that downstream effect; a matched controller-on/off factorial would be required for that causal decomposition.

\subsection{Design Implications for Trajectory Controllers}
The results distinguish \emph{answer optimization} from \emph{interface optimization}. If generated text is only a final answer for a human, unchanged aggregate accuracy and the audit ties constrain the claim. If the text is an interface to another program, explicit intermediate state is part of the system API: an omitted alternative, qualifier, or reconsideration marker can remove a branch that a deterministic controller was designed to inspect or edit.

Controller-facing prompt evaluation should therefore measure the end task, observability of downstream-consumed states, and adverse changes together. Here, accuracy measures the end task; All-8, role completion, and lexical exposure measure observable protocol state; premature commitment and the conductive contrast expose costs. The latter is especially informative because duplication improves the target SSNHL branch while weakening a nearby contrastive branch.

Placement matters too. A trailing duplicate is the strongest tested intervention, but after-query placement alone is not a strong lexical lever, so the factorial result does not reduce to recency. A practical rule is to test duplication against the controller's own endpoint and retain an adverse-effect check. A matched follow-up crossing placement with controller presence could then test interaction and mediation directly; the current study establishes the control variable and systems interaction, not that causal path.

\Needspace{9\baselineskip}
\section{Limitations and Ethics}
The seven models are a selected contemporary panel, not a random sample from a defined model population, and all three datasets are medical multiple-choice benchmarks. Transfer to other domains, instruction families, and serving stacks is unestablished. A confirmatory study should freeze the measurement stack before generation and test semantically equivalent instruction paraphrases.

Several protocol metrics are close to saturation after one instruction (7.94/8 non-trivial sections, 7.82/8 deterministic roles, 97.1\% contrastive discussion). The All-8 conjunction has more headroom because any marginal role failure makes the whole response fail; the human audit shows that many such threshold crossings are perceptually small.

TF--IDF measures lexical exposure rather than understanding and misses some synonyms, paraphrases, polarity, and orthographic variants. Length-adjusted regressions condition on a post-treatment variable and are descriptive robustness checks, not causal direct effects. All-8 is a thresholded conjunction; its relative failure reduction does not enlarge the +2.95-point absolute effect.

The blinded challenge audit used one non-clinician author and oversampled machine-positive changes, so it is neither a prevalence sample nor an inter-rater study; its 28/30 criterion was not met. The AE evidence is a downstream supporting experiment rather than a mediation test: the present design does not establish that TF--IDF, All-8 completion, or any other individual metric causes the AE gain.

The public benchmarks contain no identifiable patient data used by this study. No external human participants were recruited; human validation was an author audit of generated text. These results are not evidence of clinical safety or medical competence.

\Needspace{9\baselineskip}
\subsection{Reproducibility}
Model outputs were frozen before post-generation refinement of deterministic measurement. A stricter Step-6 semantic counterfactual classifier was dropped after independent validation; final analyses use the simpler observable criterion. Analyses are exploratory rather than preregistered confirmation; no generations were regenerated and no human ratings changed.

The original AE notebook did not retain transient case-level objects, but the reported values were checked against an independent persisted rerun containing all 4,000 responses, the exact 2,000 questions, and scoring/runtime artifacts. Code and reproduction are at the public Answer Engineering repository, \repo{}. Source and frozen outputs are bound to GitHub Release \release{}, including the 16,800-cell primary run, the complete 4,000-response AE placement rerun with question text, and integrity hashes. The AE placement experiment is also available in the public Colab notebook \colab{}. Hosted regeneration may differ because provider-side infrastructure and bitwise determinism are not fully exposed.

Verification has three levels. \emph{Frozen-output verification} checks hashes, row counts, endpoints, aggregate values, and paired transitions; \emph{deterministic rejudging} recomputes judgments from frozen responses and published scoring code; \emph{model regeneration} reruns generation. The first two audit the reported run offline; the third tests replication under a potentially changed serving stack.

\Needspace{5\baselineskip}
The AE artifact preserves exact question text for both conditions in all 2,000 paired cases, so the McNemar transitions---not only marginal percentages---remain auditable. The source tag and SHA-256 manifest bind the data to a specific code state. Paper-facing verification is offline and deterministic; fresh hosted-model generation is a separate replication exercise.

\Needspace{12\baselineskip}
\section{Conclusion}
Instruction duplication is a placement-sensitive inference-time control over machine-addressable trajectory state. In the pooled one-to-two-copy comparison, the deterministic All-8 diagnostic rises from 90.22\% to 93.17\%, pre-commitment TF--IDF exposure rises from 73.44\% to 74.81\%, and aggregate answer accuracy remains 60.21\%. A trailing copy is the strongest tested placement, while premature commitment also rises from 1.52\% to 2.30\%, exposing a real tradeoff. Many formal changes are perceptual ties to a human reader, and the prespecified audit threshold is not met.

The downstream result explains why that distinction can still matter. In the SSNHL scope, the published reason-first no-editing endpoint of 25.1\% rose to 83.5\% with AE; the later AE reproduction scored 84.2\%, and duplication raised it to 97.1\%. The conductive control moved from 58.9\% without editing to 77.9\% with published AE, 78.6\% in the reproduction, and 73.8\% with duplication---lower than system-only AE but still above the no-editing baseline. For systems that monitor or edit trajectories, copy count and placement are therefore actionable control parameters whose value should be evaluated together with the controller that consumes the trajectory.

\Needspace{8\baselineskip}
\section{Acknowledgments}
OpenAI GPT-5.5 and GPT-5.6 assisted with research-code preparation and manuscript editing. The author reviewed the resulting code, analyses, claims, and final text and takes responsibility for them.


\begin{thebibliography}{99}\footnotesize
\setlength{\bibsep}{0pt}

\bibitem[Bowen(2006)]{bowen2006reasoning}
J. L. Bowen. 2006. Educational strategies to promote clinical diagnostic reasoning. \emph{New England Journal of Medicine}, 355(21):2217--2225.

\bibitem[Charlin et al.(2000)]{charlin2000scripts}
B. Charlin, J. Tardif, and H. P. A. Boshuizen. 2000. Scripts and medical diagnostic knowledge: Theory and applications for clinical reasoning instruction and research. \emph{Academic Medicine}, 75(2):182--190.

\bibitem[Geng et al.(2023)]{geng2023grammar}
S. Geng, M. Josifoski, M. Peyrard, and R. West. 2023. Grammar-constrained decoding for structured NLP tasks without finetuning. In \emph{Proceedings of EMNLP 2023}.

\bibitem[Holm(1979)]{holm1979}
S. Holm. 1979. A simple sequentially rejective multiple test procedure. \emph{Scandinavian Journal of Statistics}, 6(2):65--70.

\bibitem[Jin et al.(2021)]{jin2021medqa}
D. Jin, E. Pan, N. Oufattole, W.-H. Weng, H. Fang, and P. Szolovits. 2021. What disease does this patient have? A large-scale open domain question answering dataset from medical exams. \emph{Applied Sciences}, 11(14):6421.

\bibitem[Kassirer(1983)]{kassirer1983hypothesis}
J. P. Kassirer. 1983. Teaching clinical medicine by iterative hypothesis testing---Let's preach what we practice. \emph{New England Journal of Medicine}, 309(15):921--923.

\bibitem[Lavrenko and Molodnitskaia(2026)]{lavrenko2026ae}
V. Lavrenko and A. Molodnitskaia. 2026. Answer Engineering: Local trajectory editing for protocol-constrained decision making in large language models. \emph{arXiv preprint} \href{https://arxiv.org/abs/2606.21121}{arXiv:2606.21121}.

\bibitem[Leviathan et al.(2025)]{leviathan2025prompt}
Y. Leviathan, M. Kalman, and Y. Matias. 2025. Prompt repetition improves non-reasoning LLMs. \emph{arXiv preprint arXiv:2512.14982}.

\bibitem[Lightman et al.(2024)]{lightman2024verify}
H. Lightman, V. Kosaraju, Y. Burda, et al. 2024. Let's verify step by step. In \emph{International Conference on Learning Representations}.

\bibitem[Liu et al.(2024)]{liu2024position}
Y. Liu, X. Zeng, F. Meng, and J. Zhou. 2024. Instruction position matters in sequence generation with large language models. In \emph{Findings of ACL 2024}, 11997--12009.

\bibitem[Lou et al.(2023)]{lou2023survey}
R. Lou, K. Zhang, and W. Yin. 2023. Large language model instruction following: A survey of progresses and challenges. \emph{arXiv preprint arXiv:2303.10475}.

\bibitem[McDuff et al.(2025)]{mcduff2025ddx}
D. McDuff, M. Schaekermann, T. Tu, et al. 2025. Towards accurate differential diagnosis with large language models. \emph{Nature}, 642:451--457.

\bibitem[Nimo et al.(2025)]{nimo2025afrimed}
C. Nimo, T. Olatunji, A. T. Owodunni, et al. 2025. AfriMed-QA: A pan-African, multi-specialty, medical question-answering benchmark dataset. In \emph{Proceedings of the 63rd Annual Meeting of the Association for Computational Linguistics (Volume 1: Long Papers)}, 1948--1973.

\bibitem[Ouyang et al.(2022)]{ouyang2022training}
L. Ouyang, J. Wu, X. Jiang, et al. 2022. Training language models to follow instructions with human feedback. In \emph{Advances in Neural Information Processing Systems}.

\bibitem[Salton and Buckley(1988)]{salton1988term}
G. Salton and C. Buckley. 1988. Term-weighting approaches in automatic text retrieval. \emph{Information Processing \& Management}, 24(5):513--523.

\bibitem[Stolfo et al.(2025)]{stolfo2025steering}
A. Stolfo, V. Balachandran, S. Yousefi, E. Horvitz, and B. Nushi. 2025. Improving instruction-following in language models through activation steering. In \emph{International Conference on Learning Representations}.

\bibitem[Turpin et al.(2023)]{turpin2023unfaithful}
M. Turpin, J. Michael, E. Perez, and S. R. Bowman. 2023. Language models don't always say what they think: Unfaithful explanations in chain-of-thought prompting. In \emph{Advances in Neural Information Processing Systems}.

\bibitem[Wei et al.(2022)]{wei2022cot}
J. Wei, X. Wang, D. Schuurmans, et al. 2022. Chain-of-thought prompting elicits reasoning in large language models. In \emph{Advances in Neural Information Processing Systems}.

\bibitem[Xu et al.(2023)]{xu2023rereading}
X. Xu, C. Tao, T. Shen, et al. 2023. Re-reading improves reasoning in large language models. \emph{arXiv preprint arXiv:2309.06275}.

\bibitem[Zhou et al.(2023)]{zhou2023ifeval}
J. Zhou, T. Lu, S. Mishra, et al. 2023. Instruction-following evaluation for large language models. \emph{arXiv preprint arXiv:2311.07911}.

\bibitem[Zuo et al.(2025)]{zuo2025medxpertqa}
Y. Zuo, S. Qu, Y. Li, Z.-R. Chen, X. Zhu, E. Hua, K. Zhang, N. Ding, and B. Zhou. 2025. MedXpertQA: Benchmarking expert-level medical reasoning and understanding. In \emph{Proceedings of the 42nd International Conference on Machine Learning}, PMLR 267:80961--80990.

\end{thebibliography}
\end{document}